\documentclass[conference]{IEEEtran}
\IEEEoverridecommandlockouts
\usepackage{cite}
\usepackage{amsmath,amssymb,amsfonts}
\usepackage{algorithmic}
\usepackage{graphicx}
\usepackage{xcolor}
\usepackage{threeparttable}
\usepackage{makecell}
\newcommand{\ignore}[1]{}

\usepackage[utf8]{inputenc}

\title{Algorithm Selection for Image Quality Assessment\\
\thanks{Presented at the Seventh Workshop on COnfiguration and SElection of ALgorithms (COSEAL), Potsdam, Germany, August 26--27, 2019.}
\thanks{Funded by the Deutsche Forschungsgemeinschaft (DFG, German Research Foundation) – Project-ID 251654672 – TRR 161}
}

\author{\IEEEauthorblockN{Markus Wagner}
\IEEEauthorblockA{\textit{University of Adelaide} \\
Adelaide, Australia \\
markus.wagner@adelaide.edu.au}
\and
\IEEEauthorblockN{Hanhe Lin}
\IEEEauthorblockA{\textit{University of Konstanz} \\
Konstanz, Germany \\
hanhe.lin@uni-konstanz.de}
\and
\IEEEauthorblockN{Shujun Li}
\IEEEauthorblockA{\textit{University of Kent} \\
Canterbury, Kent, UK \\
S.J.Li@kent.ac.uk}
\and
\IEEEauthorblockN{Dietmar Saupe}
\IEEEauthorblockA{\textit{University of Konstanz} \\
Konstanz, Germany \\
dietmar.saupe@uni-konstanz.de}
}

\date{March 2019}

\begin{document}
\maketitle
\begin{abstract}
Subjective perceptual image quality can be assessed in lab studies by human observers. Objective image quality assessment (IQA) refers to algorithms for estimation of the mean subjective quality ratings. Many such methods have been proposed, both for blind IQA in which no original reference image is available as well as for the full-reference case. We compared 8 state-of-the-art algorithms for blind IQA and showed that an oracle, able to predict the best performing method for any given input image, yields a hybrid method that could outperform even the best single existing method by a large margin. In this contribution we address the research question whether established methods to learn such an oracle can improve blind IQA. We applied AutoFolio, a state-of-the-art system that trains an algorithm selector to choose a well-performing algorithm for a given instance. We also trained deep neural networks to predict the best method. Our results did not give a positive answer, algorithm selection did not yield a significant improvement over the single best method. Looking into the results in depth, we observed that the noise in images may have played a role in why our trained classifiers could not predict the oracle. This motivates the consideration of noisiness in IQA methods, a property that has so far not been observed and that opens up several interesting new research questions and applications.
\end{abstract}


\section{Introduction}

The perceptual quality of visual media is of relevance for the development of media compression and enhancement algorithms as well as for content providers wishing to ensure sufficient user satisfaction. Assessment of visual quality requires human judges or algorithmic (``objective'') methods. These can be trained on subjective mean opinion scores (MOS) from benchmarks achieved by human lab assessments, or, more recently, by larger crowdsourcing studies. 


In this contribution we consider the case of blind image quality assessment (BIQA), i.e., the estimation of subjective perceptual image quality without availability of a pristine reference image. There are several image quality datasets available for training and testing BIQA methods, and a number of algorithms have been proposed to solve the BIQA task providing more or less accuracy. It can be expected that there is no single method that achieves the best result, i.e., a quality estimation nearest to the MOS, for all of the images in a test set. Therefore, here we consider learning to predict for each input image the best suited IQA method out of a portfolio of a set of candidate algorithms. 

This is an instance of the general algorithm selection problem \cite{rice1976algorithm}: Given a portfolio 
${\mathcal {P}}$ of algorithms or methods, a set $\mathcal {I}$ of problems,  and a cost metric $ m:\mathcal{P} \times \mathcal {I}\to \mathbb {R}$, the algorithm selection problem consists of finding a mapping $s:\mathcal {I}\to \mathcal {P}$ from instances in 
${\mathcal {I}}$ to algorithms in 
${\mathcal {P}}$ such that the total cost 
${\sum_{I\in {\mathcal {I}}}m(s(I),I)}$ across all instances is minimized. If ${\mathcal {I}}$ and 
${\mathcal {P}}$ are finite, the single best method (SBM) is given by $M^{\star}\in \mathcal {P}$ with $M^{\star} = \arg \min_{M\in \mathcal {P}} \sum_{I\in {\mathcal {I}}} m(M,I)$, and the virtual best selection model (VBM), also called the oracle, $\mathcal {O}$, is the one that selects the best algorithm in each case, so $\mathcal {O}(I) = \arg \min_{M\in \mathcal {P}} m(M,I)$ for all $I \in \mathcal {I}$.

In the case of BIQA, the set of algorithms is finite, $M_k \in \mathcal {P}, k=1, \ldots, K$, and the instances are images from a test set of a benchmark dataset, $I_n \in {\mathcal {I}, n=1,\ldots, N}$, where the image qualities have been assessed by mean opinion scores, $\text{MOS}(I_n)$. The cost function can be, e.g., the absolute error of the BIQA method, $s(I,M)=|M(I) - \text{MOS}(I)|$.

 It turns out that for a set of state-of-the-art BIQA algorithms and a large-scale image quality dataset there is a very large performance gap between the single and the virtual best method. Thus, in this work we are pursuing the research question for BIQA, whether advanced methods of algorithm selection are able to close this gap. All our attempts, however, failed in this regard. It seems, algorithm selection for BIQA does not yield an improvement over the single best method. Although negative, this result gives rise to a number of interesting new research questions, posed at the end. 

We are not aware of any previous work on algorithm selection for blind IQA as well as for the full reference case (FR-IQA). However, hybrid method have been proposed for FR-IQA, combining all methods from a portfolio by linear combinations trained by regression for the quality assessment. In addition, images can be classified according to distortion type and for each of these a separate method fusion can be designed \cite{xu2015metrics}. It was also proposed for FR-IQA to select and linearly combine a subset of IQA methods, however, globally, i.e., not adaptively for each input image  \cite{oszust2016decision}.

\section{Data set and the virtual best method}
In \cite{lin2018towards} the authors introduced a diverse dataset, called KonQ-10k, of 10,073 natural images with authentic distortions intended for machine learning BIQA methods. 
Currently, it is the largest such dataset available. It is subdivided into a training set and a test set of 8,058 and 2,015 images, respectively. 
Seven well-known IQA methods (BIQI, BLIINDS-II, etc.) and a newly developed deep learning method (KonCept512)  were applied to the test set and gave results, summarized in Table \ref{tab:methods}. More details and the references for the methods can be found in \cite{lin2018towards}. The second column of the table lists the number of features used for each method.

We have fitted the predictions of the eight methods to the ground truth values of the training set, which were scaled to the interval $[0,100]$, by nonlinear regression, using the 5-parameter logistic function from \cite{sheikh2006statistical}. This is a necessary preprocessing step before algorithm selection, because IQA methods generally are trained to give the best correlation with ground truth rather than minimizing an average error measure. In Table \ref{tab:methods} we list the Spearman rank order correlation coefficient (SROCC) and the mean absolute error (MAE) of the predictions of all methods. The MAE is based on the joint quality scale $[0,100]$.

\setlength{\tabcolsep}{1.3mm}
\renewcommand\arraystretch{1.05}
\begin{table}[t]
\caption{Performance of 8 IQA methods on the KonIQ-10k test set.}\label{tab:methods}\centering%
\begin{threeparttable}
\begin{tabular}{|l r|cc|rrr|}
  \hline
  &\multicolumn{3}{c|}{} & \multicolumn{3}{c|}{Method best for images} \\
  Method & Features & SROCC & MAE & Rank 1 & Rank 2 & Rank 3\\ \hline \hline
  BIQI       &     18 & 0.559 & 8.339 & 187 & 188 & 240\\
  BLIINDS-II &     24 & 0.585 & 9.239 & 185 & 215 & 205 \\
  BRISQUE    &     36 & 0.705 & 8.224 & 176 & 205 & 253\\
  CORNIA     & 20,000 & 0.780 & 7.308 & 217 & 263 & 286\\
  DIIVINE    &     88 & 0.589 & 8.180 & 169 & 198 & 259 \\
  HOSA       & 14,700 & \bf 0.805 & \bf 6.792 & \bf 220 & 324 & 316  \\
  SSEQ       &     12 & 0.604 & 9.403 & 179 & 227 & 168 \\
  KonCept512 & 1,536 & \bf 0.921 & \bf 4.154 & \bf 682 & 395 & 288 \\
  \hline  \hline
  Virtual best method & NA & \bf 0.978 & \bf 2.069 & \bf 2,015 & 0 & 0\\ 
  \hline
\end{tabular}
\end{threeparttable}
\end{table}

\begin{figure}[t]
\centering
\vspace{-2mm}
\includegraphics[width=1.0\columnwidth,trim={0 0 0 40},clip]{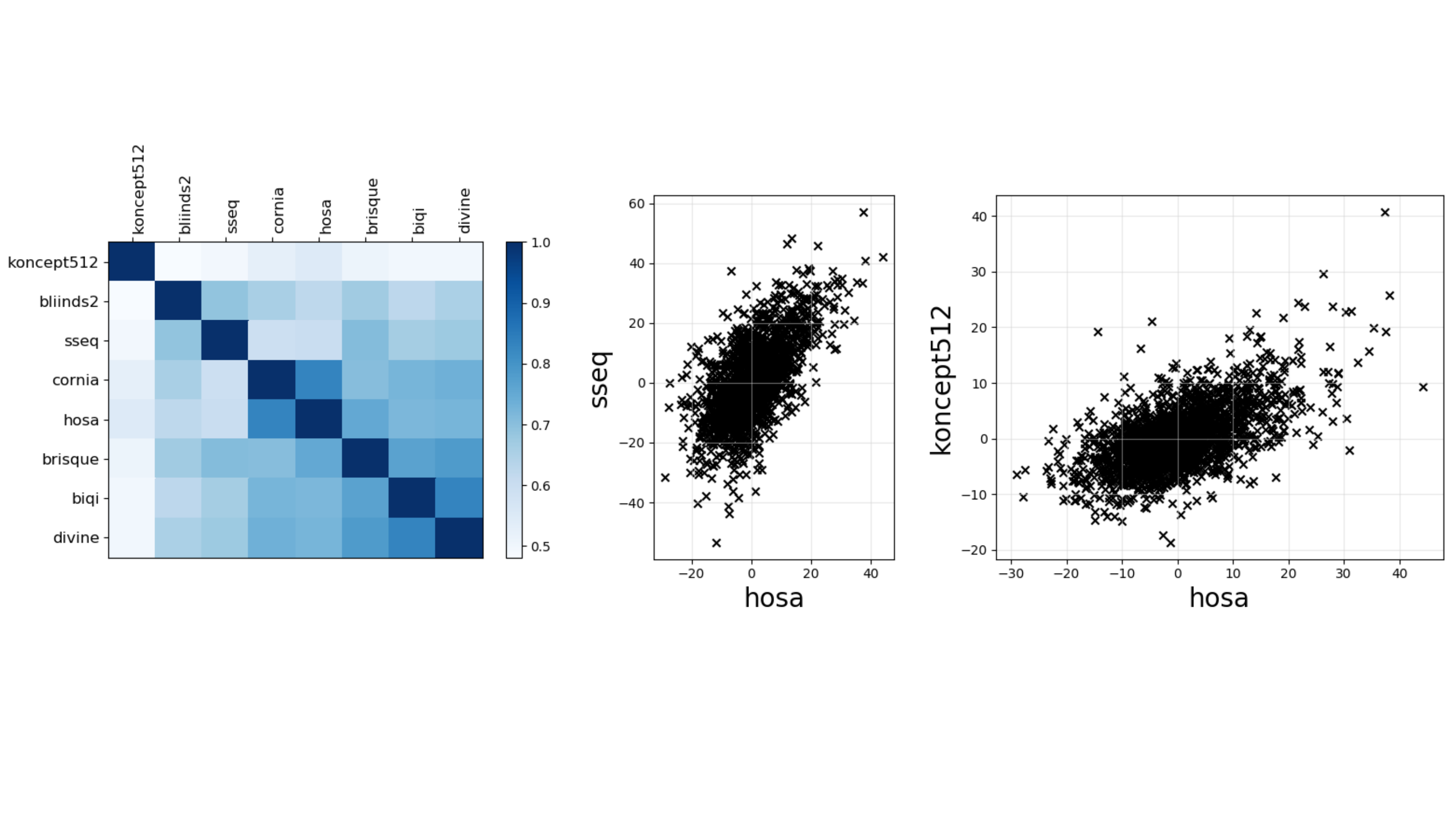}
\vspace{-20mm}
\caption{Left: The correlations (SROCC) between the predictions of the 8 selected methods, clustered with Ward's hierarchical method, are shown color coded. KonCept512's performance across all test instances is the most different from the other seven, while others are more related, such as CORNIA and HOSA, and BIQI and DIIVINE. 
Right: Two scatter plots showing the (signed) errors $M(I) - \text{MOS}(I)$ for two pairs of methods. Points clustered along the vertical axis imply that the method plotted on the horizontal axis has smaller errors, and vice versa. So HOSA is more accurate than SSEQ, but less than KonCept512. Figures were generated using ASAPy \cite{bischl-aij16a}.}
\label{fig:corr_methods}
\end{figure}

After this alignment, we obtained the virtual best method by checking for each of the 2015 test images which method estimated its quality closest to the ground truth MOS. The columns labeled ``Rank 1, 2, 3'' in Table \ref{tab:methods} show the numbers of images for which each method gave the best, the second and the third best result. The single best method, KonCept512, provided 682 out of 2,015 scores for the virtual best method. This is more than three times as many as any other method, but still only 33.8\% of all test images.
The virtual best method gave an SROCC value of 0.978 and an MAE of 2.069, much better than the single best method. 
The correlation diagram and scatter plots in Figure~\ref{fig:corr_methods}
show a certain degree of \textit{complementarity} of the algorithms, which, in principle, should allow us to train an effective algorithm selector.

\section{Algorithm selection using AutoFolio}

In our first attempt of algorithm selection for BIQA, we employed AutoFolio \cite{lindauer15autofolio}. This tool automatically determines a well-performing algorithm selection approach and its hyper-parameters. In its learning phase, AutoFolio takes as input two matrices: one that lists for each training instance its instance feature values, and the other one lists for each instance the performance of all (eight) algorithms. AutoFolio takes these and then explores the ``algorithm selector design space'', which includes design parameters such as different models (e.g., random forests and XGBoost) with various parameterizations, and preprocessing options (e.g., PCA on/off and scaling on/off).

From Table~\ref{tab:methods}, the total number of features is 36,414, mostly because CORNIA, HOSA, and KonCept512 make use of many features. To limit a possible selection bias of features by AutoFolio and to reduce complexity, we performed a principle component analysis for the set of features of each of the three methods mentioned above and then selected the most important 100 features in each case. In total, this resulted in 478 features that the eight methods contribute.

\begin{table}[t]
\caption{Performance of single best method (SBM), virtual best method (VBM), and algorithm selection (AS) by AutoFolio on the KonIQ-10k test set. The first table part shows the number of instances covered by each method. }\label{tab:results}\centering
\newcommand{\na}{\phantom{000}--}
\begin{threeparttable}
\begin{tabular}{|l|ccc|ccc|}
  \hline
  &\multicolumn{3}{c|}{Using all methods} & \multicolumn{3}{c|}{KonCept512 excluded} \\
  Method & SBM & VBM & AS & SBM & VBM & AS \\ \hline \hline
  BIQI       &  \na & 187 & \phantom{000}0 &  \na  & 263 & \phantom{0}51\\
  BLIINDS-II &  \na & 185 & \phantom{000}0 &  \na  & 277 & \phantom{0}32\\
  BRISQUE    &  \na & 176 & \phantom{000}0 &  \na  & 256 & 140\\
  CORNIA     &  \na & 217 & \phantom{000}0 &  \na  & 329 & 512\\
  DIIVINE    &  \na & 169 & \phantom{000}0 &  \na  & 241 & 252 \\
  HOSA       &  \na & 220 & \phantom{000}0 &  2015 & 363 & 918  \\
  SSEQ       &  \na & 179 & \phantom{000}0 &  \na  & 286 & 110 \\
  KonCept512 & 2015 & 682 & 2015 &  \na  & \phantom{00}-- & \phantom{00}-- \\
  \hline  \hline
  MAE   & 4.154 & 2.069 & 4.154 & 6.792 & 3.063 & 6.665\\ 
  SROCC & 0.921 & 0.978 & 0.921 & 0.805 & 0.954 & 0.784  \\ 
  \hline
\end{tabular}
\end{threeparttable}
\end{table}

Table~\ref{tab:results} lists the results of two experiments.
In both, AutoFolio was given 24 hours to explore the model space. 
In the first one, we allowed it to use all eight algorithms. Despite our and AutoFolio's best efforts (it explored over 500 models in 24 hours), the best algorithm selector chose KonCept512 for \textit{all} of the 2015 test instances, even though the VBM would pick it in just about 34\% of all cases. 
Due to KonCept512's dominance, we excluded it in the second experiment. The MAE of the remaining seven method's VBM increased to 3.063. Interestingly, AutoFolio now managed to learn an algorithm selector that performed slightly better than the single best method of the remaining seven algorithms. However, this holds only for the MAE performance metric, and not for SROCC. Moreover, the large gap to the VBM's performance (i.e., when considering just these seven) has remained.

\section{Algorithm selection using deep learning}

\begin{figure}[!t]
\centering
\includegraphics[width=0.48\textwidth]{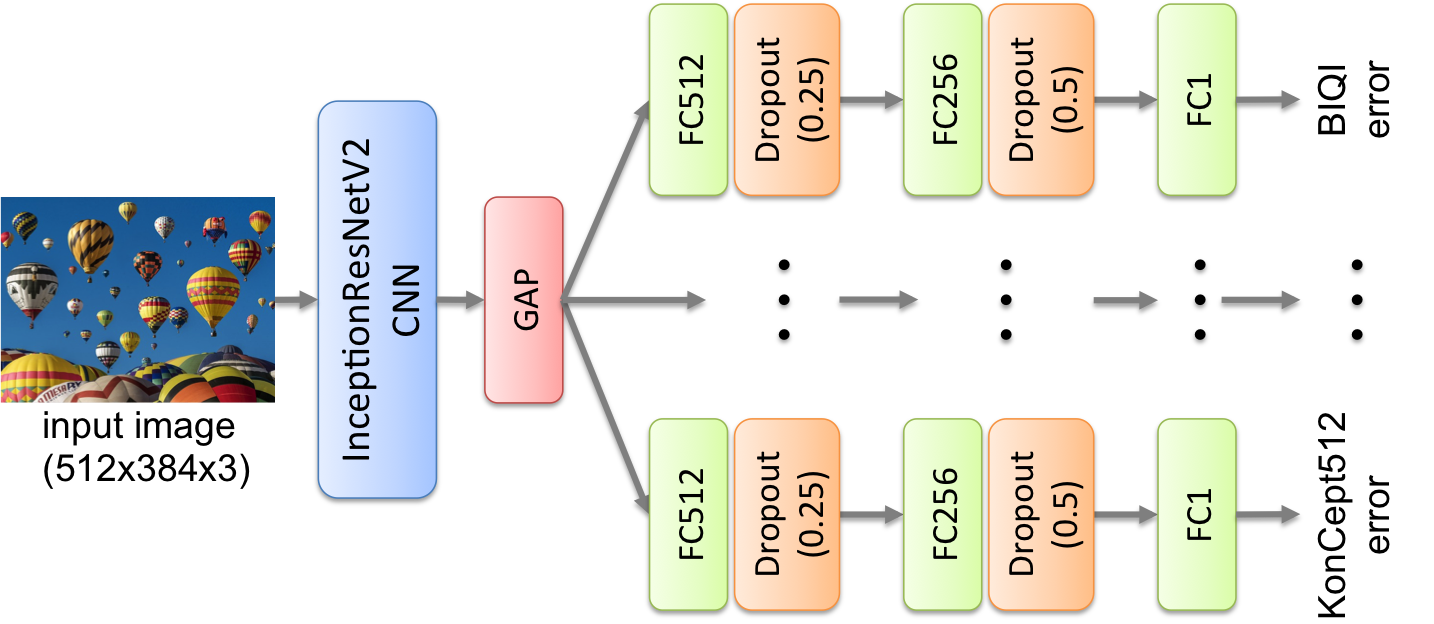}
\caption{The proposed siamese network architecture for error prediction.}
\label{fig:arch_pred_ae.png}
\end{figure}

For our second attempt, a set of images with ground truth image quality values were used and split into a subset of training images and a smaller one for validation. We tried two approaches using deep learning classifiers.

Approach 1: Train a CNN-based deep learning system to classify images according to which IQA method achieves the best image quality prediction. Thus, we consider eight classes, one for each method. After training, this network provides a solution to the algorithm selection problem. We have used InceptionResNetV2 \cite{szegedy2017inception} for the image classification problem.
Given the training set and a validation set (1,000 images, split from the training set), we fine-tuned InceptionResNetV2 with the pre-trained weight on ImageNet dataset \cite{deng2009imagenet}, where the stochastic gradient descent optimizer was applied with a small learning rate $\alpha = 0.0001$. We trained 10 epochs with a batch size of 64 and reported the model that gave the best prediction results on validation set. This gave an image classification accuracy of 29.3\% and an SROCC of 0.871 on the KonIQ-10k test set after algorithm selection.

Approach 2: For IQA methods $M$ and images $I$, having ground truth quality values $\mathrm{MOS}(I)$, we consider the error function $f_M(I) = |M(I)-\mathrm{MOS}(I)|$. We tried to learn these functions by a Siamese neural network with a joint CNN base for all considered methods $M$. 
Then the algorithm selection for a given input image would first run this network and then output the IQA $M(I)$ of the method $M$, for which the network predicted the smallest error $f_M(I)$. The proposed architecture is shown in Fig.~\ref{fig:arch_pred_ae.png}. We feed an image into the CNN base of InceptionResNetV2 and use Global Average Pooling (GAP) for each feature map. The resulting feature vector passes through 8 separate modules, each one predicting the error for one of the eight methods. Each module consists of five layers. These are of type fully-connected (FC) with 512 units, dropout with rate 0.25, FC with 256 units, dropout with rate 0.5, and output with one neuron. We replaced the cross entropy loss by mean absolute error loss and applied the same training process as in Approach~1. The model that gave the lowest loss on the validation set was accepted. For an input image $I$ it produces estimates $\hat{f}_M(I)$ of the error $f_M(I)$ for all methods $M$, leading to the algorithm selection result $M^{\star}(I) = \min_{M\in \mathcal {P}} \hat{f}_M(I)$. The MAE $f_{M^{\star}}(I)$ on the test set was 6.447, which gave an SROCC of 0.908. 


\section{Discussion and Conclusion}
The virtual best algorithm by means of algorithm selection from a portfolio of eight methods would yield an extreme improvement of IQA performance over the single best one, KonCept512 (SROCC of 0.978 versus 0.921). However, all our attempts to apply methods of algorithm selection have failed to achieve a performance better than that of the single best one. Using state-of-the-art algorithm selection, the best model came out to be equal to the best single method, KonCept512. Moreover, both approaches to learning to identify the best IQA method for an input image by deep neural networks gave results on the test set that are worse than those of the single best method (SROCCs of 0.871 and 0.908).


Our explanation is a combination of two issues. Firstly, we conjecture that the performance of the single best algorithm, KonCept512, is already close to being \textit{optimal}, i.e., at the saturation limit of what can be achieved for blind IQA on our training and test sets. Secondly, we conjecture that the clear superiority of the virtual best algorithm may be attributed to `noisy' evaluation of image quality. Consider an IQA method and a fixed test image. For this image there are numerous other images that are perceptually indistinguishable but different in terms of pixel RGB values. When evaluating an IQA method on this set of visually equivalent images, we would obtain a distribution of image quality values. So the actual quality estimate of a particular image can be interpreted as the mean value of all of these measurements, plus an added noise term. In this case the virtual best method can still achieve an improvement over the optimal method, but only due to exploitation of noise which, of course, cannot be predicted by any machine learning on a training set.

Therefore, our work, although providing a negative answer to the initial question of whether algorithm selection can improve blind image quality assessment, opens up a number of interesting new research questions: Can one quantitatively and reliably assess the noisiness of IQA methods? Does denoising of IQA methods improve their performance? And finally, does denoising remove the large gap between the single best method and the virtual best, and are denoised IQA methods better suited for the algorithm selection strategy?

\bibliographystyle{IEEEtran}
\bibliography{refs}

\begin{thebibliography}{1}
\providecommand{\url}[1]{#1}
\csname url@samestyle\endcsname
\providecommand{\newblock}{\relax}
\providecommand{\bibinfo}[2]{#2}
\providecommand{\BIBentrySTDinterwordspacing}{\spaceskip=0pt\relax}
\providecommand{\BIBentryALTinterwordstretchfactor}{4}
\providecommand{\BIBentryALTinterwordspacing}{\spaceskip=\fontdimen2\font plus
\BIBentryALTinterwordstretchfactor\fontdimen3\font minus
  \fontdimen4\font\relax}
\providecommand{\BIBforeignlanguage}[2]{{%
\expandafter\ifx\csname l@#1\endcsname\relax
\typeout{** WARNING: IEEEtran.bst: No hyphenation pattern has been}%
\typeout{** loaded for the language `#1'. Using the pattern for}%
\typeout{** the default language instead.}%
\else
\language=\csname l@#1\endcsname
\fi
#2}}
\providecommand{\BIBdecl}{\relax}
\BIBdecl

\bibitem{rice1976algorithm}
J.~R. Rice, ``The algorithm selection problem,'' in \emph{Advances in
  Computers}.\hskip 1em plus 0.5em minus 0.4em\relax Elsevier, 1976, vol.~15,
  pp. 65--118.

\bibitem{xu2015metrics}
L.~Xu, W.~Lin, and C.-C.~J. Kuo, ``Metrics fusion,'' in \emph{Visual Quality
  Assessment by Machine Learning}, ser. SpringerBriefs in Electrical and
  Computer Engineering.\hskip 1em plus 0.5em minus 0.4em\relax Springer
  Singapore, 2015, ch.~5, pp. 93--122.

\bibitem{oszust2016decision}
M.~Oszust, ``Decision fusion for image quality assessment using an optimization
  approach,'' \emph{IEEE Signal Processing Letters}, vol.~23, no.~1, pp.
  65--69, 2016.

\bibitem{lin2018towards}
H.~Lin, V.~Hosu, and D.~Saupe, ``{KonIQ-10K}: {T}owards an ecologically valid
  and large-scale {IQA} database,'' \emph{arXiv:1803.08489 (cs.CV)}, 2018.

\bibitem{sheikh2006statistical}
H.~R. Sheikh, M.~F. Sabir, and A.~C. Bovik, ``A statistical evaluation of
  recent full reference image quality assessment algorithms,'' \emph{IEEE
  Transactions on Image Processing}, vol.~15, no.~11, pp. 3440--3451, 2006.

\bibitem{bischl-aij16a}
B.~Bischl, P.~Kerschke, L.~Kotthoff, M.~Lindauer, Y.~Malitsky, A.~Frech\'ette,
  H.~Hoos, F.~Hutter, K.~Leyton-Brown, K.~Tierney, and J.~Vanschoren, ``Aslib:
  A benchmark library for algorithm selection,'' \emph{Artificial Intelligence
  Journal (AIJ)}, vol. 237, pp. 41--58, 2016.

\bibitem{lindauer15autofolio}
M.~Lindauer, H.~Hoos, F.~Hutter, and T.~Schaub, ``Autofolio: An automatically
  configured algorithm selector,'' \emph{Journal of Artificial Intelligence
  Research}, vol.~53, pp. 745--778, 2015.

\bibitem{szegedy2017inception}
C.~Szegedy, S.~Ioffe, V.~Vanhoucke, and A.~A. Alemi, ``Inception-v4,
  {I}nception-{R}es{N}et and the impact of residual connections on learning.''
  in \emph{AAAI Conference on Artificial Intelligence (AAAI)}, vol.~4, 2017,
  p.~12.

\bibitem{deng2009imagenet}
J.~Deng, W.~Dong, R.~Socher, L.-J. Li, K.~Li, and L.~Fei-Fei, ``Image{N}et: A
  large-scale hierarchical image database,'' in \emph{IEEE Conference on
  Computer Vision and Pattern Recognition (CVPR)}, 2009, pp. 248--255.

\end{thebibliography}
\end{document}